\documentclass[10pt]{article}
\PassOptionsToPackage{dvipsnames,svgnames,x11names}{xcolor}
\usepackage[citestyle=authoryear]{colab}

\newcommand{\atsign}{\textrm{@}}

\definecolor{traceT}{RGB}{245,130,32}
\definecolor{traceR}{RGB}{215,55,90}
\definecolor{traceA}{RGB}{140,55,175}
\definecolor{traceC}{RGB}{60,60,165}
\definecolor{traceE}{RGB}{0,140,150}
\newcommand{\trT}[1]{\textcolor{traceT}{#1}}
\newcommand{\trR}[1]{\textcolor{traceR}{#1}}
\newcommand{\trA}[1]{\textcolor{traceA}{#1}}
\newcommand{\trC}[1]{\textcolor{traceC}{#1}}
\newcommand{\trE}[1]{\textcolor{traceE}{#1}}
\hypersetup{pdftitle={TRACE: Temporal Retrieval with Anchored and Convergent Evidence for Long-Horizon Video Understanding}}

\title{\trT{T}\trR{R}\trA{A}\trC{C}\trE{E}: \trT{T}emporal \trR{R}etrieval with \trA{A}nchored and \trC{C}onvergent \trE{E}vidence for Long-Horizon Video Understanding}

\author{Pengyiang~Liu\,$^{1,3,*}$ \quad Junbo~Niu\,$^{2,3,*}$ \quad Xiaoyang~Hu\,$^{1}$ \quad Zhongyue~Shi\,$^{1}$ \\[0.1em]
Zitian~Wang\,$^{1}$ \quad Linjiang~Huang\,$^{1}$ \quad Si~Liu\,$^{1,\dagger}$}

\labname{Colab}
\institution{$^{1}$Colab,~Beihang~University \quad $^{2}$Peking~University \quad $^{3}$Shanghai~AI~Laboratory \\[0.15em]
$^{*}$Equal contribution. \quad $^{\dagger}$Corresponding author.}
\colabdate{EMNLP 2026}
\paperurl{https://buaa-colalab.github.io/TRACE/}
\githuburl{}
\huggingfaceurl{}
\dataurl{}

\colabrunningtitle{TRACE: Temporal Retrieval with Anchored and Convergent Evidence}

\newcommand{\tracefrontfigure}{%
\begin{center}
\begin{minipage}{\textwidth}
\centering
\includegraphics[width=0.9\textwidth]{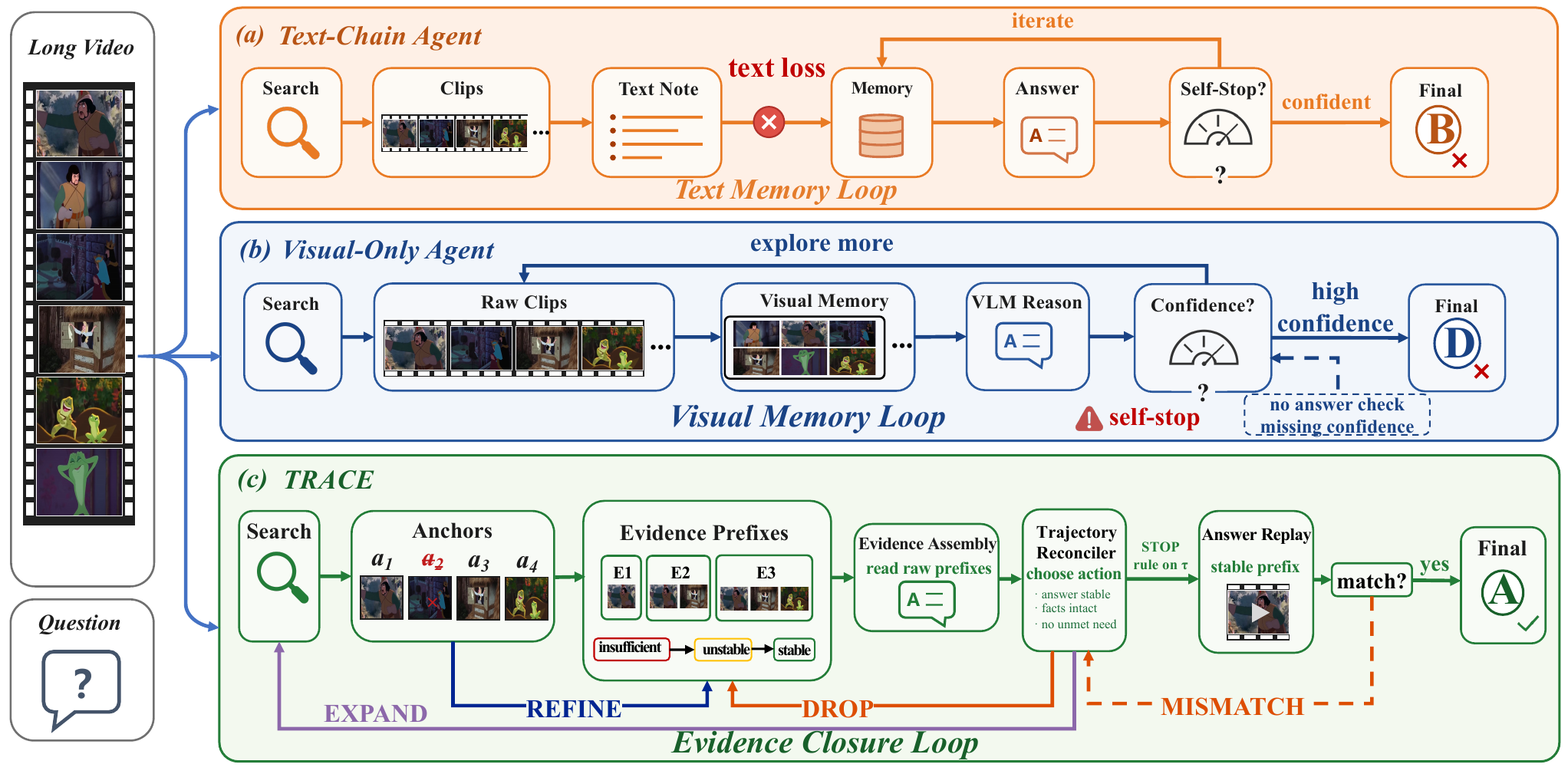}
\captionof{figure}{Comparison between text-chain agents, visual-only agents, and TRACE. TRACE keeps the answer path grounded in raw visual evidence while using trajectory reconciliation only for stopping and control.}
\label{fig:trace_comparison}
\end{minipage}
\end{center}
}

\begin{document}

\maketitle

\vspace{-12pt}

\makeatletter
\begin{colababstract}
A long-video answer is evidence-supported only when the frames decoded from the video cover every event the answer depends on. Existing evaluations score final-answer correctness or predicted evidence intervals, but the frames a method decodes before answering are rarely audited, so correct answers can still rest on incomplete observation. We introduce VES-Bench, a $600$-question benchmark of Temporal Ordering and Event Counting items over $348$ public long videos. Each item carries a jointly necessary set of evidence intervals, letting us audit at three strictness levels whether a method's decoded frames cover every one of them. We also propose TRACE, a training-free agent that grounds answers in raw visual clips, builds an evidence bundle round by round, and stops only when the answer stabilises as the bundle grows and a final pass over the same clips returns the same answer. Under a same-backbone audit, TRACE answers $50.7\%$ of questions correctly with at least two decoded frames inside every evidence interval, at $98.7$ frames per question: over $10$ points above uniform decoding at $128$ frames ($40.2\%$), and within $2.6$ points of uniform decoding at $256$ frames at $0.39\!\times$ its frame cost, while reaching the highest answer accuracy in the audit ($63.5\%$). TRACE also stays competitive on Video-MME ($86.1$), LVBench ($75.6$), and LongVideoBench ($75.1$).

\vspace{0.6em}
{\small\textbf{Project Page:}~\url{\@paperurl}}
\end{colababstract}
\makeatother

\vspace{0.5em}
\tracefrontfigure
\vspace{0.5em}

\section{Introduction}

Long-Horizon Video Understanding requires locating and integrating visual evidence that, for many questions, is distributed across multiple intervals of the video rather than concentrated in a single moment~\citep{egoschema,longvideobench,lvbench,mlvu,ovosbench}. Determining the order of several events, or counting how many times an event occurs, can only be answered from visual evidence once every relevant interval has been seen~\citep{herbench,svcbench,vrbench,videommelogical}. If any of them is left out, the remaining choices cannot be told apart on visual grounds and the model has to guess. Locating one relevant moment therefore does not imply that the model has observed everything the answer needs.

Methods for long-video question answering differ in how they accumulate evidence and where to look. Fixed-sampling and long-context approaches commit to the visible content before reasoning, so any interval missed by the initial schedule cannot be recovered later~\citep{longvila,longvu,videoxl,moviechat,malmm,goldfish,adaptivekeyframesampling}. Agent-based methods iteratively search for relevant clips~\citep{videoagent,vca,avp,lenswalk,videotree,videolucy,videoexplorer,revise}.

Figure~\ref{fig:trace_comparison} depicts two recurring patterns in agent-based methods. Text-chain agents~\citep{videoagent} accumulate evidence as a textual rationale across rounds and lose visual grounding, while visual-only agents~\citep{vca} preserve raw clips but stop on a self-reported confidence check. A wrong stop either leaves required intervals unseen or wastes the budget on confounding intervals. TRACE instead preserves visual fidelity by retaining raw visual evidence, and terminates exploration with a verifiable stopping criterion based on answer stability as the evidence bundle grows.

Existing long-horizon evaluations leave this decision unaudited. They score final-answer correctness~\citep{egoschema,videomme,longvideobench,lvbench,mlvu} or predicted evidence intervals~\citep{charadessta,momentdetr,univtg,nextgqa,deveqa,cgbench}, but not the frames a method actually decoded before answering, so a model that observed every necessary interval and one that guessed from a partial view can record the same accuracy. We close this gap with VES-Bench, a Long-Horizon Video Understanding benchmark of $600$ four-choice Temporal-Ordering and Event-Counting questions, each paired with a jointly necessary evidence set. The audit reports at three strictness levels whether the frames a method actually decoded cover that set, making evidence-supported correctness a directly measurable property of every prediction.

On VES-Bench under a shared backbone, existing agents reveal complementary blind spots. One targets evidence intervals without covering them all, while another spends a larger budget without targeting the question. TRACE combines targeting with density at an agent-comparable frame budget, surpassing agent baselines and approaching dense uniform decoding at a fraction of its frame cost. This corroborates that at a matched frame budget, the trajectory-based stopping mechanism, rather than raw budget, is what drives evidence-supported correctness. TRACE also remains competitive on Video-MME, LVBench, and LongVideoBench~\citep{videomme,lvbench,longvideobench}, without sacrificing general long-video performance.

Our contributions are as follows.
\begin{itemize}
    \item We propose TRACE, a training-free agent that grounds answers in raw visual clips and stops when the answer trajectory across the evidence bundle stabilises.
    \item We construct VES-Bench, a benchmark that audits, at three strictness levels, whether decoded frames cover each question's jointly necessary evidence intervals.
    \item At matched budgets on a shared backbone, trajectory-based stopping yields higher evidence-supported correctness than fixed-sampling and agent baselines, while TRACE stays competitive on three long-video benchmarks.
\end{itemize}

\section{Related Work}

\paragraph{Long-Horizon Video Understanding.}
Long-Horizon Video Understanding methods build on general vision-language models~\citep{nativevisual,llavaonevision2} and often address input scale by extending context length, compressing visual tokens, or constructing sparse memory. LongVILA, LongVU, and Video-XL increase the amount of video a model can process in a single pass \citep{longvila,longvu,videoxl}. MovieChat, MA-LMM, Goldfish, and InternLM-XComposer2.5-OmniLive organize long videos into compact memory or retrieval representations \citep{moviechat,malmm,goldfish,ixc25omnilive}. A complementary line of evaluation work probes whether long-video answers are visually grounded by stress-testing them with hallucination-style probes~\citep{videohallucer}. Our audit shares this motivation but operationalises grounding through observed-frame logs rather than answer-level probes.

\paragraph{Agent-based reasoning and temporal grounding.}
Two methodological lines are closely related to TRACE. Agent-based long-video frameworks formulate the task as multi-step observation, search, or tool use. VideoAgent, VideoTree, VCA, VideoMind, DeepVideoDiscovery, VideoLucy, AVP, and LensWalk explore where and how a model should observe a video, and decide when to stop \citep{videoagent,videotree,vca,videomind,deepvideodiscovery,videolucy,avp,lenswalk}. Temporal grounding methods take a complementary view, predicting query-relevant intervals end-to-end and reporting interval-level metrics such as $\mathrm{tIoU}$ and Recall@$k$ \citep{charadessta,momentdetr,univtg}. Agent and grounding lines together cover ``when and how to look'' and ``which intervals are query-relevant.'' TRACE sits between the two. As a test-time agent it controls observation through a trajectory-level stopping criterion, and VES-Bench audits, on the evaluation side, the source-frame timestamps actually decoded by each method against the annotated evidence intervals.

\paragraph{Long-video QA benchmarks.}
Long-video QA evaluation has moved beyond end-to-end answer accuracy~\citep{videomme,longvideobench,lvbench,mlvu,egoschema} toward finer diagnostics, including state-maintenance probes under streaming queries~\citep{svcbench,ovosbench,ovobench} and grounding-aware protocols. The latter typically score the evidence intervals or clues a model predicts~\citep{nextgqa,deveqa,cgbench}, or build questions that structurally require fusing cues from multiple non-overlapping clips~\citep{herbench}. VES-Bench is complementary. Rather than scoring predicted intervals or the evidential requirement of a question, it logs the source-frame timestamps actually decoded for the VLM before each final answer and checks them against per-question annotated evidence sets.

\section{Method}

\begin{figure*}[t]
\centering
\includegraphics[width=0.98\textwidth]{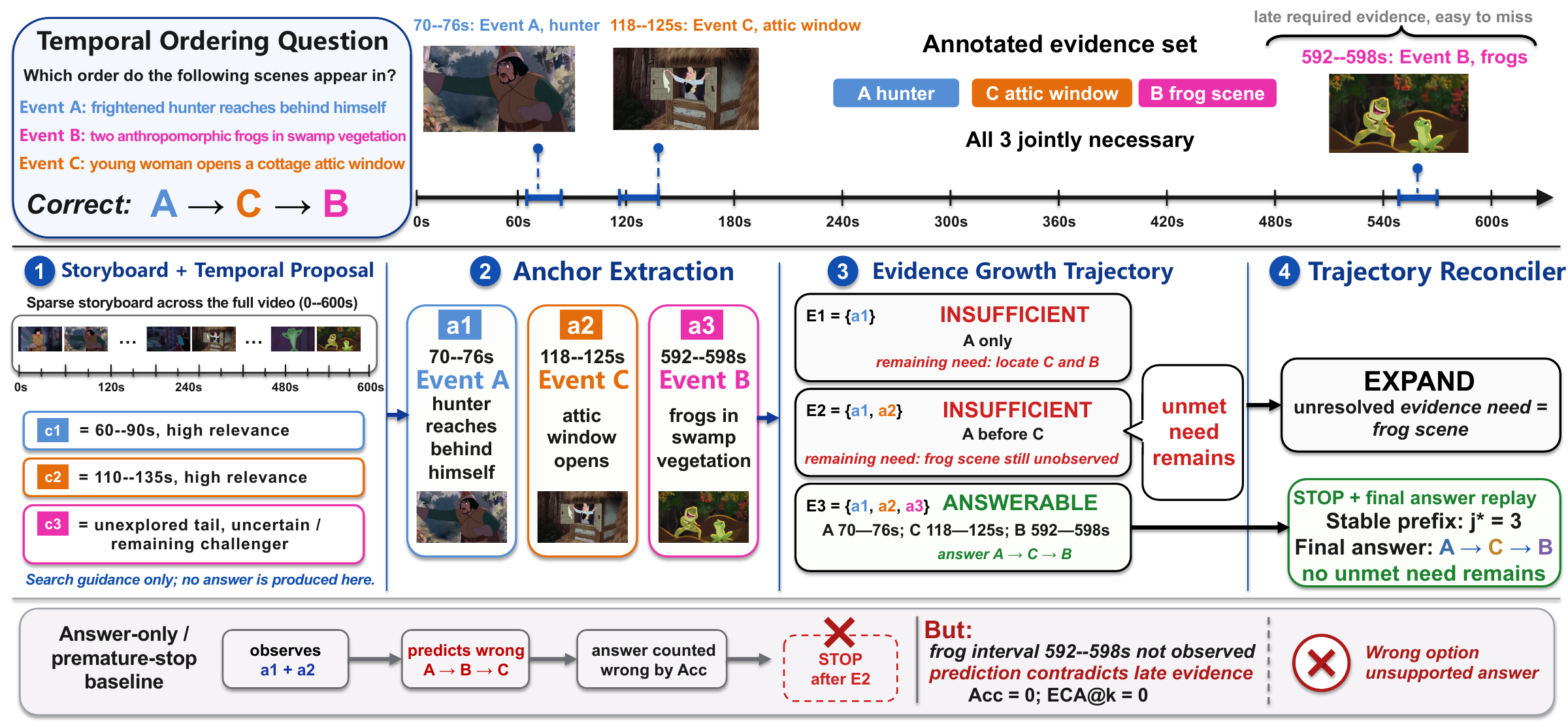}
\caption{TRACE workflow. The system converts the full video into visual anchors, constructs prefix bundles to form an evidence growth trajectory, and uses the Trajectory Reconciler to stop, drop, refine, or expand the observed evidence.}
\label{fig:trace_workflow}
\end{figure*}

TRACE organizes long-horizon video evidence as visual anchors and decides when to stop by analyzing how answers change as the evidence bundle grows. In TRACE, evidence is convergent when later visual anchors no longer change the answer, degrade answerability, contradict discriminative facts, or fill remaining evidence needs. Figure~\ref{fig:trace_workflow} summarizes the workflow, and Algorithm~\ref{alg:trace} gives the compact inference loop.

\begin{algorithm}[t]
\small
\caption{TRACE inference.}
\label{alg:trace}
\begin{algorithmic}[1]
\REQUIRE video $V$, question $q$, options $O$, budget $B$
\STATE TP, EA, and TR denote Temporal Proposal, Evidence Assembly, and Trajectory Reconciler.
\STATE Initialize storyboard $S$ as a coarse global preview of $V$, anchors $\mathcal{A}_0=\emptyset$, unexplored segments $\mathcal{R}_0$, needs $\mathcal{U}_0=\emptyset$.
\FOR{$t=1,2,\ldots$ while budget remains}
\STATE $\mathcal{P}_t \leftarrow \mathrm{TP}(S,q,O,\mathcal{R}_{t-1},\mathcal{A}_{t-1},\mathcal{U}_{t-1})$.
\STATE Extract valid anchors from $\mathcal{P}_t$; update $\mathcal{A}_t$ and $\mathcal{R}_t$.
\STATE Order anchors and form prefix bundles $\mathcal{E}_1,\ldots,\mathcal{E}_n$.
\STATE For each newly formed prefix bundle $\mathcal{E}_j$, $r_j \leftarrow \mathrm{EA}(\mathcal{E}_j,q,O)$; append to $\tau=(r_1,\ldots,r_n)$.
\STATE $(u,\mathcal{U}_t,j^\star)\leftarrow \mathrm{TR}(\tau,\mathcal{R}_t,B)$.
\IF{$u=\textsc{STOP}$ and replay$(\mathcal{E}_{j^\star})=y_{j^\star}$}
\RETURN $(y_{j^\star},\textsc{StablePrefixFound},\mathcal{E}_{j^\star})$
\ENDIF
\STATE Update state with DROP, REFINE, or EXPAND.
\ENDFOR
\STATE $P\leftarrow$ least-conflicted fallback prefix; $\hat{y}\leftarrow$ replay$(P)$.
\RETURN $(\hat{y},\textsc{NoStablePrefix},P)$
\end{algorithmic}
\end{algorithm}

\subsection{From Full Video to Visual Anchors}

A visual anchor $a$ is the basic visual evidence in TRACE: a short video clip localized within a candidate interval and defined by a temporal span with an observation strategy. The observation strategy controls frame rate and spatial resolution, allowing the system to adapt to fast motion, small objects, text, or fine-grained state changes. A visual anchor is not a textual summary, memory node, or latent representation, but raw visual evidence that Evidence Assembly can directly consume.

TRACE obtains visual anchors through three search modules. The storyboard $S$ is a coarse global preview of $V$, built once before any anchor is observed (in our implementation, $32$ uniformly spaced frames), and is not part of the prefix bundles Evidence Assembly reasons over. Like every clip subsequently supplied to Evidence Assembly, its frames are part of the actual visual input the VLM consumes on a question, so they are not a free pre-pass. At round $t$, Temporal Proposal maintains discovered candidate intervals, visual anchors, unexplored segments, and unresolved evidence needs, and consults this same preview together with the question, answer options, existing anchors, and unexplored segments to propose new candidate intervals and observation strategies. Later rounds search unexplored segments for clips that may change the answer or fill an evidence gap.

Anchor Extraction then observes each candidate interval under its assigned observation strategy and extracts a shorter visual anchor from the local video. If the interval is irrelevant to the question, no anchor is kept. Anchor Prioritization orders retained visual anchors using their visual views, the question, and the answer options, placing anchors that are more likely to distinguish choices or fill missing evidence earlier. The ordering itself does not produce an answer or final explanation. It determines how evidence is added to the evidence bundle and shapes the evidence growth trajectory.

\subsection{Constructing the Evidence Growth Trajectory}

Given ordered visual anchors $(a_{(1)},\dots,a_{(n)})$, TRACE constructs prefix bundles
\begin{equation}
\mathcal{E}_j=\{a_{(1)},\dots,a_{(j)}\},\quad j=1,\dots,n,
\end{equation}
forming a monotonic input sequence in which consecutive steps differ only by newly added visual evidence. Evidence Assembly reasons over each prefix bundle independently from raw visual clips, the question, and the answer options alone, with no carryover from earlier prefix bundles, so adjacent records $r_j$ and $r_{j+1}$ differ only in the newly added visual anchor and the trajectory $\tau$ tracks evidence growth rather than accumulated intermediate text.

Each anchor is a handful of frames, which is what keeps the trajectory affordable. Since Evidence Assembly re-supplies the whole bundle, the trajectory costs $\sum_{j}|\mathcal{E}_j|$ frames: anchors of, for example, $5$, $4$, $4$, and $3$ frames give $5+9+13+16=43$ frames across four calls. Independence between prefixes is bought with re-supplied frames, and Appendix~\ref{app:cost} reports the measured per-question total.

For each prefix bundle, Evidence Assembly outputs
\begin{equation}
r_j=(y_j,s_j,F_j,U_j),
\end{equation}
where $y_j$ is the answer candidate supported by the current evidence, $s_j$ is the answerability or conflict status, $F_j$ records discriminative facts visually supported, and $U_j$ records remaining evidence needs. All prefix-level records form the evidence growth trajectory
\begin{equation}
\tau=(r_1,\dots,r_n).
\end{equation}

TRACE differs from majority voting because the visual evidence itself changes. The question, options, prompt, and decoding setting are held fixed while only the growing visual evidence varies, so changes along $\tau$ reflect the marginal effect of new visual anchors.

\subsection{Trajectory Reconciliation and Control}

The Trajectory Reconciler analyzes the whole evidence growth trajectory and searches for a minimal stable prefix: the earliest prefix bundle from which (i) later anchors do not cause answer flips, answerability degradation, or contradiction of discriminative visual facts, and (ii) the prefix leaves no unmet evidence need. Answer-label stability alone is not evidence sufficiency. A later anchor that leaves the answer unchanged but supplies a previously missing event, identity cue, or counting fact still indicates the earlier prefix was insufficient, and a later anchor exposing conflict or poor observation quality is treated as a correction signal rather than support for stopping.

The Trajectory Reconciler emits four control actions:
\begin{itemize}\setlength{\itemsep}{0pt}\setlength{\parsep}{0pt}\setlength{\topsep}{2pt}
\item \textsc{STOP}: accept the current minimal stable prefix.
\item \textsc{DROP}: remove invalid or redundant anchors and reconstruct the trajectory.
\item \textsc{REFINE}: reobserve an existing candidate interval with a more suitable observation strategy.
\item \textsc{EXPAND}: introduce new candidate intervals from unexplored segments.
\end{itemize}
The reconciler implementation and prompt template are in Appendix~\ref{app:reconciler}, and Table~\ref{tab:reconciler_actions} lists the four actions with their triggers.

Final answer replay grounds STOP in raw visual clips. When STOP is triggered, TRACE re-answers from the raw clips in the minimal stable prefix, the question, and the answer options in a single call that returns an answer alone, and accepts STOP only if it matches $y_{j^\star}$. A mismatch becomes a conflict signal and TRACE continues with DROP, REFINE, or EXPAND while budget remains. Budget exhaustion is not treated as sufficiency. If the budget is exhausted before a reliable STOP, TRACE returns the most stable and least conflicted prefix as a forced answer without claiming sufficiency, and VES-Bench audits whether the actually observed clips support that answer.

\section{VES-Bench}

VES-Bench evaluates whether a long-video answer is supported by the visual evidence actually observed before answering. It contains $600$ four-choice questions from $348$ public Ego4D~\citep{ego4d} and YouTube videos, evenly split between Temporal Ordering and Event Counting.

\subsection{Evidence-Closed Questions}

A question is evidence-closed when it is paired with an evidence set $\mathcal{G}=\{I_1,\dots,I_m\}$ of intervals jointly necessary for determining the answer. Removing any single interval leaves at least two answer options visually compatible with the remaining evidence.

\paragraph{Temporal Ordering.}
A Temporal Ordering question asks for the relative order of several events. Its evidence set contains one interval per ordered event, and missing an event leaves its position unresolved.

\paragraph{Event Counting.}
An Event Counting question asks how many times a target event occurs. Its evidence set lists every occurrence of the target event in the video, and the count is determined only when those occurrences have been observed.

\subsection{Construction and Verification}
\label{sec:bench_construction}

VES-Bench is built by a five-stage semi-automatic pipeline. \emph{(1) Proposal}: a captioning VLM produces dense per-window captions; a proposal VLM drafts candidate items per the family schema, each carrying a question, four options, and a candidate evidence set $\mathcal{G}$ of proposed intervals. \emph{(2) Visual verification}: annotators open the video at every proposed interval, adjust temporal boundaries, and reject items whose intervals cannot be visually grounded. \emph{(3) Leave-one-out sufficiency probe}: a human annotator and an independent vision LLM both answer each $|\mathcal{G}|{-}1$ partial evidence set; an item is accepted only if both judge that no leave-one-out partial set uniquely determines the correct option. \emph{(4) Shortcut probe}: a language-only LLM answers from the question and options alone, and a separate LLM from the question, options, and a global video summary; items solvable from either are rejected. \emph{(5) Adjudication}: residual disagreements on interval boundaries, visual grounding, or leave-one-out sufficiency are resolved by a senior annotator. The candidate pool of $1{,}024$ items yields $600$ accepted items. Per-stage models, rejection rates, and inter-annotator agreement are in Appendix~\ref{app:annotation}.

\begin{figure}[t]
\centering
\includegraphics[width=0.72\columnwidth]{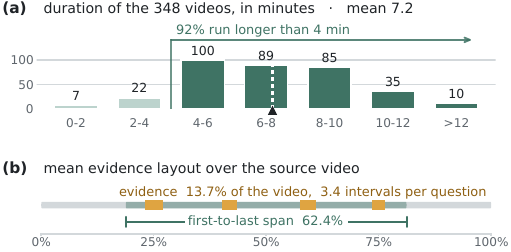}
\caption{VES-Bench duration and evidence-layout statistics.}
\label{fig:ves_statistics}
\vspace{-0.5em}
\end{figure}

\paragraph{Dataset scale.}
Figure~\ref{fig:ves_statistics} summarises the duration distribution and evidence layout. The average video lasts $434$ seconds, and an average question carries $3.4$ evidence intervals of $16.2$ seconds each, so evidence localisation over long temporal context is central to the benchmark.

\paragraph{Source and license.}
Videos are drawn from Ego4D and from public YouTube. The release carries the annotations, evidence-set boundaries, per-question observation logs, and the source videos, so that a logged timestamp can be replayed against the exact frame the audit scored. Licensing terms are in Appendix~\ref{app:licenses}.

\subsection{Audit Metrics}
\label{sec:metrics_define}

Each method receives the full video, question, and four options. The evaluator never supplies ground-truth intervals during inference. For each question $i$, the observation log $\mathcal{F}_i$ is the set of distinct source-frame timestamps decoded for the VLM before the final answer. Proposed intervals that are not decoded are not counted. We organise the audit into three measurements: targeting (EP, AR), coverage breadth (Cov@$k$), and joint correctness (ECA@$k$), all computed on $\mathcal{F}_i$ and all higher-is-better. The visual budget Fr (lower-is-better) instead sums the frames supplied over every backbone call, so a timestamp that a method re-supplies in a later call is charged again; for single-pass methods the two conventions coincide.

\paragraph{Answer accuracy.}
$\mathrm{Acc}=\tfrac{1}{N}\sum_i \mathbf{1}[\hat y_i = y_i]$.

\paragraph{Targeting (EP, EP$_{\mathrm{ref}}$, AR).}
Evidence Precision is the fraction of decoded frames that fall inside any annotated evidence interval,
\[
\mathrm{EP}=\tfrac{1}{N}\sum_i \frac{|\mathcal{F}_i\cap\bigcup_{I\in\mathcal{G}_i} I|}{|\mathcal{F}_i|}.
\]
We compare EP against a benchmark constant
\[
\mathrm{EP}_{\mathrm{ref}}=\tfrac{1}{N}\sum_i \frac{|\bigcup_{I\in\mathcal{G}_i} I|}{T_i},
\]
where $T_i$ is the source duration of the video that question $i$ is asked on, taken from the benchmark video metadata. $\mathrm{EP}_{\mathrm{ref}}$ is fixed by the benchmark itself. It is computed once from the $348$ video durations and the $600$ evidence-set annotations, and is identical for every method evaluated. On VES-Bench, $\mathrm{EP}_{\mathrm{ref}}=13.7\%$ (Appendix~\ref{app:ep_ref}). EP and $\mathrm{EP}_{\mathrm{ref}}$ share the same per-question macro-average convention, so any sampler whose frame placements are uniform in time and do not condition on question content has EP close to $\mathrm{EP}_{\mathrm{ref}}$, up to frame discretisation. We summarise targeting by the Allocation Ratio $\mathrm{AR}=\mathrm{EP}/\mathrm{EP}_{\mathrm{ref}}$, where $\mathrm{AR}=1$ corresponds to uniform allocation and $\mathrm{AR}>1$ indicates preferential observation of evidence intervals.

\paragraph{Coverage breadth (Cov@$k$).}
For each density level $k\!\in\!\{1,2,3\}$,
\[
\mathrm{Cov@}k=\tfrac{1}{N}\sum_i \prod_{I\in\mathcal{G}_i}\mathbf{1}[|\mathcal{F}_i\cap I|\ge k]
\]
is the rate at which every annotated evidence interval receives at least $k$ decoded frames. Cov@$1$ checks whether every interval has been touched at all and saturates at moderate uniform budgets. Cov@$3$ raises the bar but under-resolves the lower budgets that most agents and direct VLMs use. We use Cov@$2$ as the headline density level throughout the paper, since it is the weakest threshold that excludes single-frame coincidence inside each interval, and we report Cov@$1$ and Cov@$3$ as lower and upper sensitivity bounds. The three values together describe a graded notion of how thoroughly each evidence interval has been decoded. The headline choice does not exclude either bound.

\paragraph{Joint correctness (ECA@$k$).}
Evidence-Covered Accuracy at density $k$ couples correctness with Cov@$k$,
\[
\mathrm{ECA@}k=\tfrac{1}{N}\sum_i \mathbf{1}[\hat y_i = y_i]\prod_{I\in\mathcal{G}_i}\mathbf{1}[|\mathcal{F}_i\cap I|\ge k].
\]
ECA@$k$ is the rate of questions that are answered correctly while every evidence interval is observed at density at least $k$. By construction, $\mathrm{Acc}\!\ge\!\mathrm{ECA@}1\!\ge\!\mathrm{ECA@}2\!\ge\!\mathrm{ECA@}3$, and for any fixed Acc the spread among ECA@$k$ tracks how thoroughly the evidence supporting those correct answers has been observed. ECA@$2$ is the headline density throughout the analysis. ECA@$1$ and ECA@$3$ are reported alongside as sensitivity bounds. ECA@$k$ measures observation sufficiency; Section~\ref{sec:ves_main} pairs it with a counterfactual check that the covered intervals are the ones an answer depends on.

\paragraph{Reading the metrics.}
EP and AR describe targeting. Cov@$k$ describes coverage breadth. ECA@$k$ couples them with answer correctness. In the main audit table we pair these with the average supplied frames Fr, so each row is read as a (method, frame budget) configuration. Statistical reporting follows Appendix~\ref{app:statistics}.

\paragraph{Timestamp requirement.}
End-to-end video compression or hidden key-frame selection systems can report Acc on VES-Bench, but they can report EP, AR, Cov@$k$, and ECA@$k$ only if they expose source-frame timestamps that map back to the original video.

\section{Experiments}

We evaluate TRACE on public long-video benchmarks for answer-only competitiveness and on VES-Bench for evidence-supported correctness. The VES-Bench audit is the central study.

\subsection{Public Benchmarks}

\begin{table}[t]
\centering
\caption{Answer-only long-video benchmark accuracy (Video-MME no-subtitle; LVBench/LongVideoBench validation). Backbones in parentheses. The VCA (Gemini-2.5-Pro) rows are our reruns; all other rows reproduce the numbers reported by their sources~\citep{gemini25,qwen3vl,adaretake,qwen25vl,videolucy,deepseekr1,deepvideodiscovery,openai_o3_o4mini,vca,gpt4o,avp}. Evidence-supported claims are on VES-Bench (Table~\ref{tab:ves_main}).}
\label{tab:public}
\small
\setlength{\tabcolsep}{4pt}
\renewcommand{\arraystretch}{0.95}
\begin{tabular*}{0.92\textwidth}{@{\extracolsep{\fill}}lrrr@{}}
\toprule
Method & Video-MME & LVBench & LongVideoBench \\
\midrule
\multicolumn{4}{l}{\textit{Closed-source VLMs}} \\
Gemini-2.5-Flash & 74.2 & 62.2 & 66.2 \\
Gemini-2.5-Pro & 82.4 & 67.4 & 69.8 \\
\midrule
\multicolumn{4}{l}{\textit{Open-source VLMs}} \\
Qwen3-VL-30B-A3B & 79.2 & 67.7 & -- \\
AdaReTaKe-72B & 73.5 & 53.3 & 67.0 \\
Qwen2.5-VL-72B & 72.6 & 47.3 & 65.9 \\
\midrule
\multicolumn{4}{l}{\textit{Agent-based methods}} \\
VideoLucy (DS-R1+Qwen2.5-VL-7B) & 72.5 & 58.8 & -- \\
DeepVideoDiscovery (OpenAI o3) & -- & 74.2 & 71.6 \\
VCA (GPT-4o) & -- & 41.3 & -- \\
VCA (Gemini-2.5-Pro) & 66.0 & 52.3 & 60.7 \\
AVP (Gemini-2.5-Pro) & \underline{85.3} & \underline{74.8} & \underline{73.4} \\
\midrule
\multicolumn{4}{l}{\textit{Ours}} \\
TRACE (Gemini-2.5-Pro) & \textbf{86.1} & \textbf{75.6} & \textbf{75.1} \\
\bottomrule
\end{tabular*}
\vspace{-0.6em}
\end{table}

Table~\ref{tab:public} reports standard answer accuracy on Video-MME~\citep{videomme}, LVBench~\citep{lvbench}, and LongVideoBench~\citep{longvideobench}. Public baselines include Gemini-2.5~\citep{gemini25}, Qwen2.5-VL~\citep{qwen25vl}, Qwen3-VL~\citep{qwen3vl}, AdaReTaKe~\citep{adaretake}, VideoLucy~\citep{videolucy}, DeepVideoDiscovery~\citep{deepvideodiscovery}, VCA~\citep{vca}, and AVP~\citep{avp}. TRACE uses Gemini-2.5-Pro to match AVP's reported backbone, reaching $86.1$ on Video-MME, $75.6$ on LVBench, and $75.1$ on LongVideoBench. We treat this as a competitiveness check. The evidence-supported analysis is on VES-Bench. Backbone choice is documented in Appendix~\ref{app:backbone}.

\subsection{VES-Bench Main Results}
\label{sec:ves_main}

\begin{table*}[t]
\centering
\caption{VES-Bench audit. Random and Blind GPT-4-Turbo (text-only) are non-visual baselines; ECA\atsign$k$ is $0$ by construction. Uniform-$k$ rows report sampler statistics (Fr, EP, AR, Cov\atsign$k$) computed on the canonical centered-uniform schedule $t_j=(j+\tfrac{1}{2})T_i/n$; these are dataset-defined and identical across backbones, so the Direct VLM block at Uniform-$64$ shares them with the Gemini-3-Flash, $64$ row above and varies only in Acc and ECA\atsign$k$.}
\label{tab:ves_main}
\small
\setlength{\tabcolsep}{4pt}
\renewcommand{\arraystretch}{0.95}
\begin{tabular*}{\textwidth}{@{\extracolsep{\fill}}lcccccc@{}}
\toprule
& Fr & Acc~$\uparrow$ & EP~$\uparrow$ & AR~$\uparrow$ & Cov@$k$~$\uparrow$ & ECA@$k$~$\uparrow$ \\
Method & & (\%) & (\%) & & {\scriptsize $k\!=\!1\,/\,2\,/\,3$} & {\scriptsize $k\!=\!1\,/\,2\,/\,3$} \\
\midrule
\multicolumn{7}{l}{\itshape Non-visual baselines} \\
Random & 0 & 25.0 & -- & -- & -- & 0.0 / 0.0 / 0.0 \\
Blind GPT-4-Turbo (text-only) & 0 & 25.3 & -- & -- & -- & 0.0 / 0.0 / 0.0 \\
\midrule
\multicolumn{7}{l}{\itshape Same-backbone uniform sweep (Gemini-3-Flash)} \\
Gemini-3-Flash, 16  &  16 & 35.3 & 13.7 & 1.00 & 19.8 /  4.0 /  1.7          &  9.2 /  2.5 /  1.3 \\
Gemini-3-Flash, 32  &  32 & 42.7 & 14.0 & 1.02 & 48.0 / 13.2 /  5.2          & 20.7 /  6.7 /  3.2 \\
Gemini-3-Flash, 64  &  64 & 43.8 & 13.7 & 1.00 & 82.2 / 34.3 / 16.3          & 36.8 / 15.5 /  8.2 \\
Gemini-3-Flash, 128 & 128 & 57.5 & 13.8 & 1.01 & 97.8 / 74.7 / 45.0 & 56.0 / 40.2 / 22.3 \\
Gemini-3-Flash, 256 & 256 & 56.7 & 13.7 & 1.00 & \textbf{99.7} / \textbf{95.5} / \textbf{81.8} & 56.5 / \textbf{53.3} / \textbf{44.5} \\
\midrule
\multicolumn{7}{l}{\itshape Direct VLMs at Uniform-64} \\
Doubao-Seed-2.0     & 64 & 27.2 & -- & -- & -- & 22.2 /  8.3 /  4.5 \\
Qwen3-VL-8B         & 64 & 43.7 & -- & -- & -- & 36.3 / 14.3 /  7.5 \\
Qwen3-VL-30B-A3B    & 64 & 44.3 & -- & -- & -- & 38.5 / 15.2 /  8.0 \\
Qwen3.6-35B-A3B     & 64 & 47.5 & -- & -- & -- & 39.5 / 14.7 /  7.7 \\
Gemini-2.5-Pro      & 64 & 48.7 & -- & -- & -- & 40.2 / 15.7 /  7.2 \\
\midrule
\multicolumn{7}{l}{\itshape Same-backbone agent baselines (Gemini-3-Flash)} \\
VCA   & 34.6  & 37.8             & \underline{16.9} & \underline{1.23} & 22.0 /  5.8 /  2.0          &  8.3 /  3.3 /  1.5 \\
AVP   & 101.3 & \underline{59.0} & 13.7             & 1.00             & 95.3 / 68.2 / 43.3          & \underline{58.2} / 39.7 / 20.8 \\
TRACE &  98.7 & \textbf{63.5}    & \textbf{29.1}    & \textbf{2.12}    & \underline{98.3} / \underline{78.5} / \underline{60.3} & \textbf{63.0} / \underline{50.7} / \underline{39.0} \\
\bottomrule
\end{tabular*}
\vspace{-0.6em}
\end{table*}

All VES-Bench experiments in this section use Gemini-3-Flash~\citep{gemini_models} as the shared backbone, including the direct VLM at multiple frame budgets, AVP~\citep{avp}, VCA~\citep{vca}, and TRACE. The direct VLM block additionally reports Doubao-Seed-2.0~\citep{seed20}, two Qwen3-VL variants~\citep{qwen3vl}, Qwen3.6-35B-A3B~\citep{qwen36}, and Gemini-2.5-Pro~\citep{gemini25} to span answerer capability. Holding the backbone fixed isolates the effect of observation policy at a comparable per-question frame budget (Appendices~\ref{app:reproduction} and~\ref{app:backbone}).

\paragraph{Direct VLMs at $64$ frames sit at the targeting reference.}
Under Uniform-$64$ the canonical sampler yields $\mathrm{EP}=13.7$ at $\mathrm{EP}_{\mathrm{ref}}$ ($\mathrm{AR}=1.00$) and $\mathrm{Cov}@\{1,2,3\}=82.2/34.3/16.3$, identical across all six backbones evaluated at this budget. Acc therefore reflects answerer capability alone, sitting above the $25.3$ language-only anchor (Blind GPT-4-Turbo~\citep{gpt4}, Appendix~\ref{app:blind_llm}). Doubao-Seed-2.0 sits at $27.2$, near the four-option chance rate, and marks the lower end of the capability range this block spans. ECA@$k$ additionally carries the Cov@$k$ constraint.

\paragraph{Same-backbone uniform sweep on Gemini-3-Flash.}
Holding the backbone fixed and sweeping supplied frames from $16$ to $256$ traces a frame--ECA frontier. ECA@$2$ moves $2.5\!\to\!6.7\!\to\!15.5\!\to\!40.2\!\to\!53.3$ at $16/32/64/128/256$ frames. The $128\!\to\!256$ step lowers Acc by $0.8$ points ($57.5\!\to\!56.7$) while Cov@$2$ gains $20.8$ ($74.7\!\to\!95.5$), so beyond a certain budget additional frames strengthen coverage but not accuracy. AR stays near $1.0$ and EP near $\mathrm{EP}_{\mathrm{ref}}$ across the sweep, since uniform decoding cannot condition on question content.

\paragraph{Same-backbone agent baselines.}
VCA spends $34.6$ frames with $\mathrm{AR}=1.23$ and $\mathrm{EP}=16.9$, the only same-backbone agent baseline with AR above $1$, but its coverage stays narrow (Cov@$1=22.0$, Cov@$2=5.8$, ECA@$2=3.3$). AVP spends $101.3$ frames with $\mathrm{AR}=1.00$ and $\mathrm{EP}=13.7$, indistinguishable from uniform allocation. Its Cov@$2=68.2$ and ECA@$2=39.7$ are comparable to same-backbone Uniform-$128$ (ECA@$2=40.2$), so under the audit AVP behaves as a uniform-density allocation at a slightly lower frame count. The two agents fail in complementary ways: VCA targets without covering, AVP covers without targeting.

\begin{figure}[t]
\centering
\includegraphics[width=0.62\columnwidth]{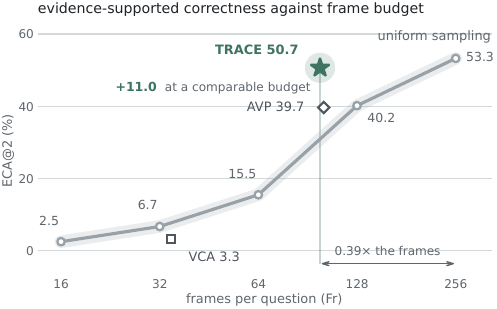}
\caption{Frame budget against evidence-supported correctness, plotted from Table~\ref{tab:ves_main} on a log frame axis. The grey curve is the same-backbone uniform sweep; each point is labelled with its ECA\atsign$2$. TRACE ($98.7$ frames) and AVP ($101.3$) sit on the vertical rule, $2.6$ frames apart.}
\label{fig:frontier}
\end{figure}

\begin{figure}[t]
\centering
\includegraphics[width=0.62\columnwidth]{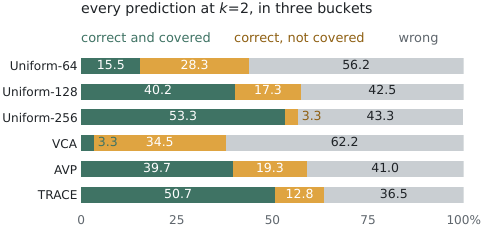}
\caption{Every prediction at $k\!=\!2$ split into Cov-Corr (correct, Cov\atsign$2$ passing), Uncov-Corr (correct without Cov\atsign$2$), and Wrong. Cov-Corr is ECA\atsign$2$ and Cov-Corr$+$Uncov-Corr is Acc, both in Table~\ref{tab:ves_main}, up to independent rounding of each value.}
\label{fig:buckets}
\end{figure}

\paragraph{TRACE shifts the frame--ECA frontier.}
TRACE operates at $98.7$ frames with $\mathrm{AR}=2.12$ and $\mathrm{EP}=29.1$, the largest departure from $\mathrm{EP}_{\mathrm{ref}}$ in the audit. Figure~\ref{fig:frontier} places the three agents against the uniform sweep. Cov@$2=78.5$ and ECA@$2=50.7$ exceed both AVP at a comparable budget and uniform decoding at $128$ frames, and fall within $2.6$ ECA@$2$ points of uniform decoding at $256$ frames at $0.39\!\times$ the frame cost. Across $k\!=\!1,2,3$, TRACE keeps the highest ECA@$k$ among same-backbone methods at comparable Fr.

Figure~\ref{fig:buckets} decomposes every prediction at $k\!=\!2$ into Cov-Corr (correct, Cov@$2$ passing), Uncov-Corr (correct without Cov@$2$), and Wrong. Direct VLMs at $64$ frames have Uncov-Corr larger than Cov-Corr, so a substantial share of correct answers rests on partial evidence. AVP and Uniform-$128$ leave Uncov-Corr at $19.3$ and $17.3$, while Uniform-$256$ and TRACE push it down to $3.3$ and $12.8$, with TRACE doing so at roughly the AVP frame budget.

A counterfactual check separates supplied evidence from used evidence. For each of the $304$ predictions TRACE answers correctly with Cov@$2$ satisfied, masking one annotated interval---its frames replaced by adjacent non-evidence frames at a fixed budget---changes the answer in $78.0\%$ of cases, while masking a random non-evidence segment covering the same number of decoded frames changes it in $6.9\%$. Coverage therefore tracks the evidence an answer depends on.

\begin{table}[t]
\centering
\caption{Per-family breakdown. ECA\atsign$k$ at $k\!=\!1,2,3$. Temporal Ordering has the largest spread across $k$ (short per-event intervals); Event Counting Acc plateaus near $40\%$ across the Gemini-3-Flash budget points, reflecting a recognition bottleneck.}
\label{tab:family_breakdown}
\small
\setlength{\tabcolsep}{3pt}
\renewcommand{\arraystretch}{0.95}
\begin{tabular*}{\textwidth}{@{\extracolsep{\fill}}lcccc@{}}
\toprule
& \multicolumn{2}{c}{Temporal Ordering} & \multicolumn{2}{c}{Event Counting} \\
\cmidrule(lr){2-3}\cmidrule(l){4-5}
Method & Acc~$\uparrow$ & ECA@$k$~$\uparrow$ & Acc~$\uparrow$ & ECA@$k$~$\uparrow$ \\
       & (\%)           & {\scriptsize $1\,/\,2\,/\,3$} & (\%)           & {\scriptsize $1\,/\,2\,/\,3$} \\
\midrule
Gemini-3-Flash, 64  & 49.7 & 37.0 /  6.3 /  1.0 & 38.0 & 36.7 / 24.7 / 15.3 \\
Gemini-3-Flash, 128 & 75.0 & 72.0 / 45.3 / 14.7 & \underline{40.0} & \underline{40.0} / 35.0 / 30.0 \\
Gemini-3-Flash, 256 & 74.3 & 74.0 / \textbf{67.7} / \textbf{52.7} & 39.0 & 39.0 / \textbf{39.0} / \underline{36.3} \\
VCA                 & 48.0 &  2.7 /  0.0 /  0.0 & 27.7 & 14.0 /  6.7 /  3.0 \\
AVP                 & \underline{78.3} & \underline{76.7} / 43.3 / 12.3 & 39.7 & 39.7 / 36.0 / 29.3 \\
TRACE               & \textbf{82.0} & \textbf{82.0} / \underline{64.3} / \underline{41.3} & \textbf{45.0} & \textbf{44.0} / \underline{37.0} / \textbf{36.7} \\
\bottomrule
\end{tabular*}
\vspace{-0.6em}
\end{table}

Per-family Table~\ref{tab:family_breakdown} confirms TRACE improves over the same-backbone agent baselines on both families, with the gain concentrated on Temporal Ordering. Event Counting Acc plateaus near $40\%$ across the Gemini-3-Flash budget points, indicating a recognition-related residual error not addressable by frame budget alone.

\subsection{Ablations}

\begin{table}[t]
\centering
\caption{Ablation of TRACE components on VES-Bench (Gemini-3-Flash). Headline densities Cov\atsign$2$, ECA\atsign$2$; full $k\!=\!1,2,3$ in Appendix Table~\ref{tab:ablation_full}.}
\label{tab:ablation}
\small
\setlength{\tabcolsep}{4pt}
\renewcommand{\arraystretch}{0.95}
\begin{tabular*}{0.90\textwidth}{@{\extracolsep{\fill}}lcccc@{}}
\toprule
Variant & Acc~$\uparrow$ & EP~$\uparrow$ & Cov@2~$\uparrow$ & ECA@2~$\uparrow$ \\
\midrule
TRACE                              & \textbf{63.5} & \textbf{29.1} & \textbf{78.5} & \textbf{50.7} \\
\;\;w/o final answer replay        & \underline{59.7} & \underline{28.2} & \underline{77.3} & \underline{42.0} \\
\;\;w/o growth trajectory          & 55.0 & 19.4 & 56.3 & 27.5 \\
\;\;w/o EXPAND                     & 53.0 & 26.4 & 49.0 & 25.5 \\
\;\;fixed-round stopping           & 52.3 & 25.7 & 47.2 & 24.0 \\
\bottomrule
\end{tabular*}
\vspace{-0.6em}
\end{table}

Table~\ref{tab:ablation} isolates the contributions of the main TRACE components. Removing the evidence growth trajectory reduces EP from $29.1$ to $19.4$ and ECA@$2$ from $50.7$ to $27.5$, while Acc decreases from $63.5$ to $55.0$. The trajectory carries the targeting behaviour that drives EP and the dense-coverage component of ECA@$k$. Fixed-round stopping and removing EXPAND retain EP between $25$ and $27$ but reduce Cov@$2$ below $50$. Removing final answer replay leaves EP and Cov@$2$ essentially unchanged but reduces ECA@$2$ by $8.7$ points, since the stable prefix is then accepted on its Evidence Assembly record alone, without a second answer-only pass over the same clips. Two components emerge: trajectory-based acquisition (EP, Cov@$k$) and final answer replay (extra ECA@$k$ on a stable prefix). See Appendix~\ref{app:trajectory_vs_voting} for the contrast with voting-based methods.

\subsection{Qualitative Analysis}

A representative Temporal Ordering item asks the chronological order of three visually distinct scenes ($A$: a frightened hunter at $70$--$76$s, $C$: a young woman opening an attic window at $118$--$125$s, $B$: two anthropomorphic frogs at $592$--$598$s). The correct order $A\!\to\!C\!\to\!B$ is unsupported unless the late frog interval is also observed, and increasing density requires that interval to be revisited substantively rather than glanced at once. TRACE handles this case by continuing acquisition while the trajectory still exposes an unresolved evidence need, and by accepting a stable prefix only after final answer replay against its raw clips.

\section{Conclusion}

We presented TRACE and VES-Bench for evidence-supported long-horizon video understanding. VES-Bench pairs answer correctness with targeting, coverage breadth, and joint correctness, audited at three strictness levels. TRACE pairs the evidence growth trajectory with final answer replay to ground its stopping decision, reaching $\mathrm{ECA@}2=50.7$ at $98.7$ frames on a same-backbone audit.

\section*{Limitations}
VES-Bench covers two evidence-closed families (Temporal Ordering, Event Counting). Extending the audit to families with looser evidence boundaries is left to future work. The headline audit uses Gemini-3-Flash because it requires per-question source-frame timestamps and full inference-loop control across every method, ablation, and budget point. We expect the qualitative ordering to hold on other backbones. TRACE depends on the backbone for visual recognition, and the Event Counting Acc plateau near $40\%$ from $64$ to $256$ frames on Gemini-3-Flash reflects a recognition bottleneck additional frames do not address. Trajectory inference adds prefix-bundle reasoning calls that scale with retained anchors, kept small by rule-triggered STOP (per-question backbone-call breakdown and frame budget in Appendix~\ref{app:cost}). Annotation depends on human judgement of evidence-set closure, which the semi-automatic pipeline (Section~\ref{sec:bench_construction}) reduces but does not remove.

\section*{Ethics Statement}
VES-Bench draws its videos from Ego4D~\citep{ego4d} and from public YouTube. The annotations, evidence-set boundaries, and per-question observation logs are released under CC BY-NC-SA 4.0 for non-commercial research, and the source videos accompany them so that the observation logs can be replayed frame by frame. Downloading the release binds the user to those terms and to the licenses of the source datasets, listed in Appendix~\ref{app:licenses}. Ego4D footage is egocentric and depicts identifiable people; the benchmark inherits the consent and de-identification terms of that dataset and adds no new recording. Annotation was carried out by four of the authors, three as annotators and one as adjudicator, over roughly $470$ person-hours. The audit metrics read decoded frame timestamps and annotated interval boundaries, so they depend on no attribute of the people appearing in the videos.

\section*{Acknowledgments}
This research is supported in part by the National Natural Science Foundation of China (No.\ 62461160308, U23B2010, 62576024), the Beijing Natural Science Foundation (No.\ L231011), the Fundamental Research Funds for the Central Universities (No.\ 501RCQD2025141003), the BeiHang GanWei Project (No.\ 502GWXM2024141001), and the National Science Foundation Support Projects (No.\ 62425303).

{\small
\bibliographystyle{\colabbibstyle}
\bibliography{references}
}

\clearpage
\appendix
\section*{Appendix}

\section*{Content of Appendices}

\noindent\small
\setlength{\tabcolsep}{4pt}
\renewcommand{\arraystretch}{0.95}
\begin{tabularx}{0.84\textwidth}{@{}p{0.18\textwidth}X@{}}
Section A. & Implementation Details and Per-question Cost \\
Section B. & Trajectory Reconciler Pipeline \\
Section C. & Trajectory vs.\ Majority Voting and Self-Consistency \\
Section D. & Same-Backbone Reproduction of Agent Baselines \\
Section E. & Statistical Reporting \\
Section F. & Targeting Reference \\
Section G. & Backbone Configuration \\
Section H. & Annotation Pipeline Details \\
Section I. & Blind LLM Validation \\
Section J. & Ablation Across Density Levels \\
Section K. & Dataset Licenses \\
\end{tabularx}
\normalsize
\vspace{0.8em}

\section{Implementation Details and Per-question Cost}
\label{app:implementation}

TRACE is implemented as prompt-based test-time modules around the same backbone used in each evaluation protocol. Temporal Proposal, Anchor Extraction, Anchor Prioritization, Evidence Assembly, and the Trajectory Reconciler exchange structured records for search and control, while final answer replay consumes only raw visual clips, the question, and answer options, and emits an answer alone. All observation accounting follows the decoded source-timestamp protocol defined in the main text.

\paragraph{Per-question cost.}
\label{app:cost}
All Section~\ref{sec:ves_main} methods run on Gemini-3-Flash via Vertex AI. Temporal Proposal consumes a $32$-frame storyboard built once per question, and later rounds reuse that preview. Evidence Assembly is invoked once per prefix bundle, the Trajectory Reconciler only when STOP is not proposed (STOP is rule-triggered from $\tau$ alone, with the backbone invoked only for DROP/REFINE/EXPAND), and one final replay reads the stable prefix. AVP under the same audit makes $\sim\!3.2$ planner calls dominated by text-only replanning over the question, defaulting to whole-video coverage when no localised cue is present---the text-chain pattern flagged in Section~1. Visual input dominates each call in our Vertex per-call records (visual\,:\,text token ratio $\sim\!100\!:\!1$, hundreds of visual tokens per decoded frame; questions, options, and trajectory/reconciler JSON together contribute under $1\%$ of input tokens), so frame budget acts as a tight proxy for per-call observation cost and we report it as the headline cost. TRACE consumes $98.7$ frames per question under the Fr convention of Section~\ref{sec:metrics_define}. Each anchor is a handful of frames, so the largest single Evidence Assembly call is the full retained set rather than the per-question total. The per-question total is in the same range as AVP's $101.3$.

\section{Trajectory Reconciler Pipeline}
\label{app:reconciler}

The Trajectory Reconciler (TR) consumes the trajectory $\tau=(r_1,\dots,r_n)$, the unexplored segments $\mathcal{R}$, and the remaining budget $B$, and emits exactly one of \textsc{STOP}, \textsc{DROP}, \textsc{REFINE}, \textsc{EXPAND}. The controller is a deterministic preprocessing step followed, when needed, by a single backbone call.

\paragraph{Deterministic preprocessing.}
TR first computes three flags from $\tau$ alone:
\begin{itemize}
\item \emph{Stable-prefix candidate} $j^\star$: the smallest $j$ such that $y_j=y_{j+1}=\dots=y_n$, $s_j$ is answerable, $F_j$ is not contradicted by any $F_{j'>j}$, and $U_j$ is empty.
\item \emph{Conflict set} $\mathcal{C}\subseteq\mathcal{A}$: anchors whose addition flips $y$ or contradicts an earlier $F$.
\item \emph{Unmet-need set} $\mathcal{U}^\star$: the union of $U_n$ and any need that recurs across the last two prefixes.
\end{itemize}
If $j^\star$ exists, TR proposes \textsc{STOP} at $j^\star$; the definition of $j^\star$ already excludes anchors after $j^\star$ that flip $y$ or contradict $F_{j^\star}$, so $\mathcal{C}$ is read only by the control call below. The proposal still has to pass final answer replay (Section~3) before being accepted as the answer. STOP is therefore rule-triggered, not LLM-decided.

\paragraph{Single backbone call for control.}
If \textsc{STOP} is not proposed, TR issues exactly one backbone call with a structured prompt containing (i) the question and answer options; (ii) $\tau$ in compact JSON form (\texttt{prefix\_id, answer, status, facts\_supported, unmet\_needs}); (iii) the conflict set $\mathcal{C}$; (iv) the unmet-need set $\mathcal{U}^\star$; (v) a list of unexplored segments with durations; and (vi) the remaining budget. The call returns one action: \textsc{DROP} naming a subset of $\mathcal{C}$, \textsc{REFINE} naming an existing anchor with a new observation strategy chosen from a fixed set (increase frame rate, raise spatial resolution, narrow temporal span, or shift temporal centre), or \textsc{EXPAND} naming up to $K$ candidate intervals from $\mathcal{R}$. The backbone is constrained to act as a classifier over a small action space, not as a free-form reasoner over the question.

\paragraph{Why a backbone call rather than rules alone.}
The deterministic flags are sufficient for STOP, but choosing among DROP/REFINE/EXPAND requires resolving which anchor in $\mathcal{C}$ to remove, which observation strategy is appropriate for a partially-grounded interval, and which unexplored segment most plausibly fills $\mathcal{U}^\star$. These are visual judgements that do not have closed-form rules.

\paragraph{Prompt template (excerpt).}
\begin{quote}\small
\texttt{System:} You are a control module that selects one observation action. Do not answer the question. Available actions: \textsc{DROP}, \textsc{REFINE}, \textsc{EXPAND}.\\
\texttt{Input:} \{question, options, trajectory:\,[\{prefix\_id, answer, status, facts, unmet\_needs\},\,...], conflict\_set, unmet\_needs, unexplored\_segments, budget\_remaining\}.\\
\texttt{Decision rules:} (1) If \texttt{conflict\_set} is non-empty, prefer \textsc{DROP} unless the conflicting anchor still contains uniquely supporting visual evidence. (2) If \texttt{unmet\_needs} can be resolved by re-observing an existing anchor with finer rate/resolution, prefer \textsc{REFINE}. (3) Otherwise, prefer \textsc{EXPAND} on the unexplored segment most likely to contain the unmet evidence; never propose more than $K$ segments.\\
\texttt{Output (JSON):} \{action, target\_anchors or target\_intervals, observation\_strategy?\}.
\end{quote}

\begin{table}[!t]
\centering
\caption{Control actions emitted by the Trajectory Reconciler.}
\label{tab:reconciler_actions}
\footnotesize
\setlength{\tabcolsep}{4pt}
\renewcommand{\arraystretch}{0.95}
\begin{tabularx}{0.96\textwidth}{@{}p{0.12\textwidth}XX@{}}
\toprule
Action & Trigger & Execution \\
\midrule
STOP & A minimal stable prefix exists: later anchors change neither its answer nor its supported facts, and it has no unmet evidence need. & Select the prefix bundle for final answering. \\
DROP & Some anchors introduce noise, conflict, or misleading local evidence. & Remove target anchors and reconstruct the trajectory. \\
REFINE & An observed interval is relevant but lacks temporal or spatial detail. & Reobserve the candidate interval with a more suitable observation strategy. \\
EXPAND & The current evidence bundle lacks events, states, or identity cues that may affect the answer. & Introduce new candidate intervals from unexplored segments. \\
\bottomrule
\end{tabularx}
\vspace{-0.6em}
\end{table}

\section{Trajectory vs.\ Majority Voting and Self-Consistency}
\label{app:trajectory_vs_voting}

Majority voting and self-consistency~\citep{wang2023selfconsistency} repeatedly reason over the same input and aggregate variation induced by sampling or model uncertainty; the visual evidence presented to the model does not change. TRACE keeps the question, options, prompt, and decoding setting fixed while only the visual evidence grows along $\tau$, so changes between $r_j$ and $r_{j+1}$ reflect the marginal effect of the new visual anchor $a_{(j+1)}$ rather than sampling noise. Two consequences follow. First, an answer flip in $\tau$ is informative about evidence sufficiency, not about decoding variance, so a stable suffix of $\tau$ is a stronger stopping signal than agreement across resampled outputs. Second, $\tau$ exposes per-prefix unmet needs $U_j$ that allow EXPAND/REFINE to act on what is missing, while majority voting has no analogue of $U_j$ and therefore no mechanism to acquire new visual evidence. The ablation row \emph{w/o growth trajectory} in Table~\ref{tab:ablation} drops EP from $29.1$ to $19.4$ and ECA@$2$ from $50.7$ to $27.5$ while keeping the same prefix-bundle structure, indicating that the gain attributable to the trajectory itself is not captured by sampling-based aggregation.

\section{Same-Backbone Reproduction of Agent Baselines}
\label{app:reproduction}

For the VES-Bench audit we reproduce AVP and VCA on Gemini-3-Flash to hold visual recognition fixed across methods. We follow each method's published pipeline and replace only the backbone calls; intermediate LLMs used for planning or scoring are kept on the same backbone family to avoid mixing recognition strengths. Prompts are ported from each paper's released templates, with minor formatting adjustments for the Gemini API (system-role placement, tool-call schema). Frame-budget controllers are kept as published, and per-frame source timestamps are logged through a shared decoding wrapper that all three methods (AVP, VCA, TRACE) call, so observation logs are directly comparable. As a sanity check, the reproduced AVP at Gemini-2.5-Pro on Video-MME, LVBench, and LongVideoBench lands within $0.6$ points of the originally reported numbers, indicating no substantive deviation introduced by our pipeline. VCA publishes on GPT-4o, so Table~\ref{tab:public} lists that published row alongside our Gemini-2.5-Pro rerun.

\section{Statistical Reporting}
\label{app:statistics}

For VES-Bench, confidence intervals are computed by bootstrap resampling over the $600$ questions ($1{,}000$ resamples, percentile method) for Acc, EP, AR, Cov@$k$, ECA@$k$, and the three-bucket decomposition. Pairwise method comparisons use paired bootstrap on the same question set, and a video-cluster bootstrap is used as a robustness check for videos contributing multiple questions. Main-paper tables report point estimates to keep the layout compact; Table~\ref{tab:ci} reports $95\%$ intervals at $k\!=\!2$.

\begin{table}[!t]
\centering
\caption{$95\%$ bootstrap confidence intervals at $k\!=\!2$ (Gemini-3-Flash). Paired bootstrap shows TRACE vs.\ AVP differences in EP, Cov\atsign$2$, and ECA\atsign$2$ are significant at $p<0.01$; AR rescales EP by the constant $\mathrm{EP}_{\mathrm{ref}}$ and carries the same test.}
\label{tab:ci}
\small
\setlength{\tabcolsep}{4pt}
\renewcommand{\arraystretch}{0.95}
\begin{tabular*}{0.70\textwidth}{@{\extracolsep{\fill}}lcccc@{}}
\toprule
Method & Acc & EP & Cov@2 & ECA@2 \\
\midrule
G3F, $64$  & $43.8_{\pm 3.8}$ & $13.7_{\pm 1.1}$ & $34.3_{\pm 3.8}$ & $15.5_{\pm 2.8}$ \\
G3F, $128$ & $57.5_{\pm 3.9}$ & $13.8_{\pm 1.1}$ & $74.7_{\pm 3.2}$ & $40.2_{\pm 3.9}$ \\
G3F, $256$ & $56.7_{\pm 4.1}$ & $13.7_{\pm 1.1}$ & $95.5_{\pm 1.7}$ & $53.3_{\pm 4.2}$ \\
VCA        & $37.8_{\pm 3.9}$ & $16.9_{\pm 1.0}$ &  $5.8_{\pm 1.9}$ &  $3.3_{\pm 1.4}$ \\
AVP        & $59.0_{\pm 3.9}$ & $13.7_{\pm 0.4}$ & $68.2_{\pm 3.7}$ & $39.7_{\pm 3.9}$ \\
TRACE      & $63.5_{\pm 3.9}$ & $29.1_{\pm 1.5}$ & $78.5_{\pm 3.3}$ & $50.7_{\pm 4.0}$ \\
\bottomrule
\end{tabular*}
\vspace{-0.6em}
\end{table}

\section{Targeting Reference}
\label{app:ep_ref}

The Allocation Ratio AR in the main paper compares each method's EP against the benchmark constant $\mathrm{EP}_{\mathrm{ref}}$ defined in Section~\ref{sec:metrics_define}. The constant is computed per question as the union length of evidence intervals divided by the source-video duration, then averaged over the $600$ questions, using source durations from the benchmark video metadata. On VES-Bench, $\mathrm{EP}_{\mathrm{ref}}=13.7\%$. The constant is computed once from the dataset and is identical for every method evaluated; it is reported as a numeric anchor against which to read AR.

\section{Backbone Configuration}
\label{app:backbone}

All Gemini calls in this paper are served through Google's Vertex AI. The public answer-only benchmarks and the VES-Bench audit use different backbones because the two protocols impose different constraints. On the public benchmarks (Video-MME, LVBench, LongVideoBench), TRACE uses Gemini-2.5-Pro because the strongest reported agent baseline (AVP) is reported on this backbone and changing it would invalidate the comparison. On VES-Bench, the audit requires per-frame source timestamps and full inference-loop control on every method, ablation, and budget point. Running the full $600$-question audit on Gemini-2.5-Pro across the multi-budget uniform sweep, all same-backbone agent baselines, the TRACE main configuration, and the ablation suite is prohibitively expensive under our compute budget; we therefore use Gemini-3-Flash as the main audit backbone, the most capable Gemini variant for which we hold full inference-loop access at a controllable, affordable per-call cost. As a same-frame anchor across protocols, Table~\ref{tab:ves_main} reports Gemini-2.5-Pro at $64$ frames in the Direct VLM block, allowing the qualitative ordering across backbones to be read directly without re-running the full audit.

\section{Annotation Pipeline Details}
\label{app:annotation}

This appendix elaborates the five-stage pipeline from Section~\ref{sec:bench_construction}.

\paragraph{Models used.}
The pipeline uses models that do not overlap with the VES-Bench audit backbones (Gemini-3-Flash, Gemini-2.5-Pro, Doubao-Seed-2.0, Qwen3-VL-8B, Qwen3-VL-30B-A3B, Qwen3.6-35B-A3B) to avoid self-evaluation. Stage~1 dense per-window captioning uses Qwen2.5-VL-72B~\citep{qwen25vl}; Stage~1 item proposal uses GPT-4o~\citep{gpt4o}; the Stage~3 leave-one-out vision probe uses Claude~3.5 Sonnet~\citep{claude35sonnet}; the Stage~4 language-only shortcut probe uses GPT-4o (text-only); the Stage~4 gist-only probe uses Claude~3.5 Sonnet on the global summary alone. Captioning windows are $30$~seconds long with $5$-second overlap. Proposal prompts are conditioned on the family schema (Temporal Ordering, Event Counting) so each candidate item is structurally well-formed before any human inspection. Each Stage~1 proposal additionally pairs every interval in $\mathcal{G}$ with a per-interval rationale describing the visual fact the interval should attest; the rationale focuses Stage~2 verification and is discarded once the evidence set is finalised.

\paragraph{Per-stage rejection counts.}
The candidate pool of $1{,}024$ proposed items reduces to $\sim\!820$ after Stage~2 visual verification, $\sim\!700$ after Stage~3 leave-one-out probe, $\sim\!640$ after Stage~4 shortcut probe, and $600$ after Stage~5 adjudication. Stage~2 removes the most items overall, Stage~3 is the largest closure-specific filter, and Stage~4 removes what survives visual grounding but stays solvable from text or gist alone; the combined accept rate is $58.6\%$.

\paragraph{Annotators and time.}
Annotation is performed by four of the authors: three act as annotators and one as adjudicator. Each item is assigned to two of the three annotators independently, with the adjudicator resolving disagreements. The mean per-item time is $13.8$ minutes during visual verification and the leave-one-out probe combined; total annotator hours across all $1{,}024$ candidates are $\sim\!470$ hours.

\paragraph{Agreement.}
Inter-annotator IoU agreement on evidence-interval boundaries is $0.83$ (mean over the $600$ accepted items). Leave-one-out sufficiency agreement between human and Claude~3.5 Sonnet is $91.3\%$, and the shortcut probe rejects an item whenever either GPT-4o (language-only) or Claude~3.5 Sonnet (gist-only) selects the correct option above the chance baseline of $25\%$ on a held-out repeat.

\paragraph{Scope of the closure guarantee.}
The pipeline rejects items decidable from language priors, four-option chance, or coarse global gist, and it requires both human and an independent vision LLM to judge that no leave-one-out partial set uniquely determines the answer. It does not guarantee that the evidence set is the \emph{minimal} sufficient set; it guarantees that, under the audit, removing any single annotated interval leaves at least two options visually compatible with what the model has seen.

\section{Blind LLM Validation}
\label{app:blind_llm}

We run a Blind GPT-4-Turbo~\citep{gpt4} (text-only) on all $600$ VES-Bench items, supplying only the question stem and the four options; no frames, no captions, no global summary, and no retrieval. The model scores $25.3\%$ overall ($25.0\%$ Temporal Ordering, $25.7\%$ Event Counting), within the bootstrap interval of the $25\%$ random baseline. Stage~4 rejects items that GPT-4o answers correctly from text alone, so this check measures whether that filter transfers to a language-only model held out from construction; it does. ECA@$k$ is $0$ by construction since $\mathcal{F}_i=\emptyset$.

\section{Ablation Across Density Levels}
\label{app:ablation_full}

Table~\ref{tab:ablation_full} reports the ablation in Table~\ref{tab:ablation} across density levels $k\!=\!1,2,3$. Cov@$k$ and ECA@$k$ both fall as $k$ rises for every variant, and removing the growth trajectory costs the largest share of Cov@$k$ at $k\!=\!3$ ($60.3$ to $30.5$).

\begin{table}[!t]
\centering
\caption{Ablation of TRACE components on VES-Bench across density levels. The row for TRACE matches Table~\ref{tab:ablation}; the columns expose how each ablated component shifts Cov\atsign$k$ and ECA\atsign$k$ at $k\!=\!1,2,3$.}
\label{tab:ablation_full}
\small
\setlength{\tabcolsep}{4pt}
\renewcommand{\arraystretch}{0.95}
\begin{tabular*}{\textwidth}{@{\extracolsep{\fill}}lccc@{}}
\toprule
Variant & Acc~$\uparrow$ & Cov@$k$~$\uparrow$ & ECA@$k$~$\uparrow$ \\
        &      & {\scriptsize $1\,/\,2\,/\,3$} & {\scriptsize $1\,/\,2\,/\,3$} \\
\midrule
TRACE                              & \textbf{63.5} & \textbf{98.3} / \textbf{78.5} / \textbf{60.3} & \textbf{63.0} / \textbf{50.7} / \textbf{39.0} \\
\;\;w/o final answer replay        & \underline{59.7} & \underline{97.8} / \underline{77.3} / \underline{59.5} & \underline{58.0} / \underline{42.0} / \underline{31.0} \\
\;\;w/o growth trajectory          & 55.0 & 90.3 / 56.3 / 30.5 & 49.5 / 27.5 / 16.3 \\
\;\;w/o EXPAND                     & 53.0 & 84.2 / 49.0 / 24.8 & 44.0 / 25.5 / 13.7 \\
\;\;fixed-round stopping           & 52.3 & 82.3 / 47.2 / 23.5 & 43.5 / 24.0 / 13.0 \\
\bottomrule
\end{tabular*}
\vspace{-0.6em}
\end{table}

\section{Dataset Licenses}
\label{app:licenses}

The VES-Bench annotations---questions, options, evidence-set boundaries, and per-question observation logs---are released under CC BY-NC-SA 4.0 for non-commercial research. The source videos accompany them, because the audit is defined on decoded source-frame timestamps and reproducing it requires the exact frames those timestamps index. Downloading the release binds the user to CC BY-NC-SA 4.0 and to the licenses of the source datasets in Table~\ref{tab:licenses}.

\begin{table}[!t]
\centering
\caption{Licenses of the VES-Bench source datasets.}
\label{tab:licenses}
\small
\begin{tabular}{@{}ll@{}}
\toprule
Source & License \\
\midrule
Ego4D~\citep{ego4d}  & Ego4D License Agreement \\
Public YouTube       & Terms of the original uploader \\
\bottomrule
\end{tabular}
\vspace{-0.6em}
\end{table}

\end{document}